\def\year{2021}\relax
\documentclass[letterpaper]{article} % DO NOT CHANGE THIS
\usepackage{aaai21}  % DO NOT CHANGE THIS
\usepackage{times}  % DO NOT CHANGE THIS
\usepackage{helvet} % DO NOT CHANGE THIS
\usepackage{courier}  % DO NOT CHANGE THIS
\usepackage[hyphens]{url}  % DO NOT CHANGE THIS
\usepackage{graphicx} % DO NOT CHANGE THIS
\usepackage{natbib}  % DO NOT CHANGE THIS AND DO NOT ADD ANY OPTIONS TO IT
\usepackage{caption} % DO NOT CHANGE THIS AND DO NOT ADD ANY OPTIONS TO IT
\usepackage[switch]{lineno}  %

\nocopyright

\usepackage{times,amsmath,epsfig}

\usepackage{graphicx}%插入图片

\usepackage{amssymb}%数学符号

\usepackage{amsthm}%数学定理

\usepackage{amsmath}%数学公式、矩阵、积分求和等

\usepackage{enumitem}%项目编号

\usepackage{multirow} %多行合并
\usepackage[caption=false, font=footnotesize]{subfig}

\usepackage{array} %\调用公式宏包的命令应放在调用定理宏包命令之前，也能控制表格

\usepackage{booktabs} %调整表格线与上下内容的间隔

\usepackage{longtable}%调用跨页表格

\usepackage{bm}%数学字体加粗

\usepackage{verbatim} %多行注释
\usepackage{float}
\newcommand{\etal}{\textit{et al.}}
\newcommand{\tabincell}[2]{\begin{tabular}{@{}#1@{}}#2\end{tabular}}
\usepackage{color}

\newcolumntype{H}{>{\setbox0=\hbox\bgroup}c<{\egroup}@{}}
\title{Perceptron Synthesis Network: Rethinking the Action Scale Variances in Videos}
\author {
	Yuan Tian\textsuperscript{\rm 1},
	Guangtao Zhai\textsuperscript{\rm 1},
	Zhiyong Gao\textsuperscript{\rm 1}
	\\
}
\affiliations {
	\textsuperscript{\rm 1}Institute of Image Commu. and Network Engin., Shanghai Jiao Tong University, China
}
\begin{document}
%	\linenumbers  %
	\maketitle
	
	\begin{abstract}
		Video action recognition has been partially addressed by the CNNs stacking of fixed-size 3D kernels.
		However, these methods may under-perform for only capturing rigid spatial-temporal patterns in single-scale spaces,
		while neglecting the scale variances across different action primitives.
		To overcome this limitation, we propose to learn the optimal-scale kernels from the data.
		More specifically, an \textit{action perceptron synthesizer} is proposed to generate the kernels from a bag of fixed-size kernels that are interacted by dense routing paths.
		To guarantee the interaction richness and the information capacity of the paths, we design the novel \textit{optimized feature fusion layer}.
		This layer establishes a principled universal paradigm that suffices to cover most of current feature fusion techniques (\textit{e.g.}, channel shuffling and channel dropout) for the first time.
		By inserting the \textit{synthesizer}, our method can easily adapt the traditional 2D CNNs to the video understanding tasks such as action recognition with marginal additional computation cost.
		The proposed method is thoroughly evaluated over several challenging datasets (\textit{i.e.}, Somehting-to-Somthing, Kinetics and Diving48) that
		highly require 
		temporal reasoning or appearance discriminating,
		achieving new state-of-the-art results.
		Particularly,
		our low-resolution model outperforms the recent strong baseline methods, \textit{i.e.}, TSM and GST, with less than 30\% of their computation cost.
	\end{abstract}

	\section{Introduction}
	
	Video action recognition has draw much attentions in computer vision community for its tremendous applications~\cite{tian2019video}.
	Inspired by the breakthrough brought by CNNs on still image recognition task,
	% \textit{i.e.}, the classification performance surpassing the human on ImageNet~\cite{deng2009imagenet}, 
	recent video recognition methods also leverage the 2D CNNs expanded with temporal modeling ability, particularly the 3D CNNs, for spatial-temporal modeling.
	
	3D convolutions are the main operation in the 3D CNNs, which learn spatial-temporal filters to capture visual cues and dynamics of the objects simultaneously.
	To constrain the parameter number and the computation cost of the whole architecture, most 3D CNNs tend to utilize the filters of typical size 3$\times$3$\times$3,
	resulting in a relatively small receptive field.
	Stacking the small filters and dowsampling the features after each stage can enlarge the receptive field to cover the whole input clip.
	However, there are two issues that limit the performance of down stream tasks.
	(a) The \textit{scale} of receptive field \textit{w.r.t.} each layer is fixed, thus the actions of big objects can only be detected in the deeper layers, resulting in the loss of fine-grained details.
	(b) The spatial-temporal \textit{aspect ratio} of receptive field is also fixed, which relies on the hand-crafted tuning for datasets of different complexities.
	
	\begin{figure}
		\centering
		% include second image
		\small
		\includegraphics[width=8cm]{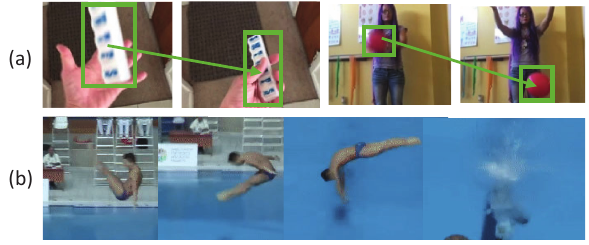} 
		\caption {
			\small
			(a) Examples from the Something-Something dataset~\cite{goyal2017something}.
			% 		The motion degrees of different actions are hugely different and the scales of involved objects are also diverse.
			The motion degree (\textit{i.e.}, the green arrow line) and the scale of moving object (\textit{i.e.}, the area of green square box) of different samples are drastically distinct from each other.
			%		This requires multi-scale spatial modeling.
			(b) Example video of “forward and PIKE” from the Diving48~\cite{li2018resound}. The action can only be recognized by first discriminating the short-term action
			primitives and then reasoning the long-term dependence order.
			%		This requires multi-scale temporal modeling.
			% 		Therefore, a unified method coping with diverse action variances in spatial-temporal space is necessary.
			Therefore, a powerful method that suffices to handle diverse action variances (in both of spatial and temporal spaces) is critical.
		}
		\label{fig_1}
	\end{figure}
	
	Particularly, the spatial-temporal scale variation is an essential property of the natural actions in the wild.
	As shown in Fig.~\ref{fig_1}, in single frame, some objects occupy a rather large spatial spaces (\textit{e.g.}, humans)
	while some other objects (\textit{e.g.}, socket and ball) only take up a relatively smaller spatial region.
	% while other objects can have a relative smaller representations (\textit{e.g.}, the socket and the ball).
	The issue becomes more serious when considering the object variance along the temporal axis:
	the same object may show different scales across different frames for the deformation of itself or the changing of the camera viewpoint.
	Moreover, tolerating the actions of different complexity is also another challenge.
	For the simple actions composed of one or two stages, such as that from the Something-Something dataset~\cite{goyal2017something}, the actions can be classified accurately by simply observing the short-term state changes within 2 or 3 frames.
	Conversely, the complex actions in Diving48~\cite{li2018resound} require the long-term comprehensive understanding of the phased changes.
	To capture such a diversity of object spatial scales and action durations,
	%even the primitives of event,
	utilizing a \textit{single} and \textit{fixed} type of kernel with a rigid spatial-temporal receptive field can not handle complex scale variances and thus results in performance degradation. 
	% may not be an optimal solution to cope with the complexity.
	For image recognition, one of the most popular architecture for multi-scale modeling is the Inception networks~\cite{szegedy2016rethinking}.
	% which extracts the multi scale features by fixed kernels of diverse sizes, \textit{e.g.},
	%1$\times$1, 3$\times$3 and 5$\times$5.
	However, simply extending this architecture to its 3D version maybe sub-optimal because the extended spatial-temporal cube-shaped kernels,
	\textit{e.g.}, 3$\times$3$\times$3 and 5$\times$5$\times$5, are not able to reach the best performance, as shown in the ablation studies.
	
	So, \textbf{how to decide the optimal-size kernels?}
	Our answer is to learn it from the data.
	Instead of simply utilizing the NAS (neural architecture search)~\cite{zoph2016neural} as in~\cite{piergiovanni2019tiny}, which involves training hundreds of models with huge computation cost,
	we seek for a soft fusion of the candidate fixed-size kernels with only \textit{one-time} training.
	More specifically, we propose an \textit{action perceptron synthesizer} to continuously generate rich-scale filters from the candidates under the optimization goal: towards the best accuracy for action classification.
	When the process finished, we freeze the synthesizer and utilize the produced filter group as the optimal.
	To fully exploit the inter-relations between the fixed-size filters and retain the network capacity to the utmost extent,
	we propose a novel \textit{optimized feature fusion layer} as a component of the synthesizer, which provides the dense learnable routing paths among the filters.
	%From the perspective of data streaming, it provide the paths for information to flow through exhausted combinations of spatial filters and temporal filters.
	%From the perspective of network architecture, it can be seen as a dynamic filter, which generates spatial-temporal filters of diverse shapes for different input videos.
	This layer covers most feature fusion techniques as special cases of it, \textit{e.g.}, channel shuffling~\cite{zhang2018shufflenet} and channel dropout.
	In the experiment part, we demonstrate that the proposed layer achieves conspicuous performance improvement while only introducing marginal learnable parameters and computation cost compared with other conventional feature fusion methods.
	
	Unlike the previous works utilizing single-scale kernels, the synthesizer allows us to deeply exploit the adaptive interactions within the multi-scale spatial and temporal information in each layer when trained on different datasets.
	By analyzing the statistical distribution of the produced kernels, we demonstrate a series of \textit{interpretable insights} in the experiment part, which are useful for the hand-crafted designing of future video networks. 
	
	%Through our extensive experiments over several large-scale action recognition datasets, \textit{i.e.}, Something-Something v1~\cite{goyal2017something}, Kinetics~\cite{carreira2017quo} and Diving48~\cite{li2018resound}, we demonstrate that our method compares favorably with widely-used 3D CNNs and other light-weight temporal reasoning modules designed upon the 2D CNNs for video modeling, and achieves state-of-the-art results.
	We summarize our contributions as follows:
	\begin{itemize}
		\setlength{\itemsep}{0pt}
		\setlength{\parsep}{0pt}
		\setlength{\parskip}{0pt}
		\item To cope with the essential spatial-temporal scale variances in the videos, we propose a novel \textit{action perceptron synthesizer} to generate the optimal-size kernels in each layer instead of simply leveraging the ordinal single-scale kernels of fixed-size.
		
		\item We propose an \textit{optimized feature fusion layer} as a component of the synthesizer
		to facilitate the inter-connections between the features from different branches, which outperforms other feature fusion methods conspicuously with negligible parameters and costs.
		
		\item We perform an extensive ablation analysis of the proposed method and also show some network designing insights from the searching process of the optimal kernel.
		
		\item We achieve state-of-the-art on several large scale video datasets with comparable parameters and FLOPs compared to existing approaches.
	\end{itemize}

	\section{Related Works}
	\subsection{Deep Video Recognition}
	%Early works tend to fine-tune the 2D CNN networks on video dataset while relying on the optical flow input modality or the subsequent temporal reasoning modules to capture the dynamic information.
	%~\cite{simonyan2014two}\cite{feichtenhofer2016convolutional} takes in the RGB input (spatial stream) and the optical flow input (temporal stream) respectively.
	Simonyan \etal~\cite{simonyan2014two} first proposed the Two-stream framework.
	Feichtenhofer \etal~\cite{feichtenhofer2016convolutional} then improved it.
	Later, TSN~\cite{wang2018temporal} proposes a new sparse frame sampling strategy.
	% and uses temporal consensus functions to aggregate features of each frame.
	%Temporal relation network (TRN)~\cite{zhou2018temporal} learns temporal dependencies between video frames at multiple time scales.
	%These methods achieve favorable results on the small video datasets, \textit{e.g.}, UCF101~\cite{soomro2012ucf101} and HMDB51~\cite{kuehne2011hmdb}.
	%However, the limited parameter number and the separated temporal modeling of the methods above may hinder the performances on large scale datasets, \textit{e.g.}, Kinetics~\cite{carreira2017quo}.
	%To learn the temporal evolution along with the spatial information simultaneously, 
	3D networks, \textit{e.g.}, C3D network~\cite{tran2015learning}, I3D~\cite{carreira2017quo}, 3D-ResNet~\cite{hara2017learning,tian2020self}, R(2+1)D CNNs~\cite{tran2018closer}\cite{qiu2017learning} and Slowfast networks~\cite{feichtenhofer2019slowfast}, recently have gained much attention as another research line.
	
	%C3D~\cite{tran2015learning} network is the first 3D network with only few layers. However, it has a huge number of parameters and is relatively hard to train.
	%I3D~\cite{carreira2017quo}, which is abbreviated from inflated 3D networks, proposes to inflate the 2D CNNs~\cite{szegedy2017inception}\cite{he2016deep} pretrained on ImageNet~\cite{deng2009imagenet} by copying weights and achieves promising results on Kinetics~\cite{carreira2017quo} dataset.
	%3D-ResNet~\cite{hara2017learning} systematically evaluates several popular inflated structures and finds that inflated 2D-ResNet inherits the advantages of ResNet~\cite{he2016deep}, i.e., easy to converge and consume low computation cost.
	%R(2+1)D~\cite{tran2018closer} decomposes the 3D convolution into a 2D convolution followed by 1D convolution and learns discriminative enough features for action recognition.
	%Slowfast networks~\cite{feichtenhofer2019slowfast} involve a slow pathway to capture spatial semantics at low frame rate and a fast pathway to capture motion at fine temporal resolution.
	%We value the merits of modeling spatial and temporal features separately.
	
	\subsection{Multi-scale CNN Architectures}
	CNNs are naturally equipped with multi scale feature representation ability due to the hierarchical stacking of convolution kernels, \textit{e.g.}, VGGNet~\cite{simonyan2014very} and ResNet~\cite{he2016deep}.
	Modern CNN architectures design the multi-scale branches explicitly.
	The GoogLeNet~\cite{szegedy2015going} utilizes diverse filters with different kernel sizes in parallel to encode the multi-scale feature in each branch.
	%However, the feature diversity is often limited by the computational constraints due to its limited parameter efficiency.
	Latter, the Inception Nets~\cite{szegedy2016rethinking}\cite{szegedy2017inception} propose to utilize more small filters in each branch of the parallel branches in the GoogLeNet~\cite{szegedy2015going} to further expand the receptive field.
	
	%Segmentation problems usually rely on large receptive field to capture the global contexts for finer results.
	%Max pooling and dilation convolution (also named atrous convolution)~\cite{yu2015multi} are widely adopted upon the backbone networks for this goal.
	%Pyramid scene parsing network (PSPNet)~\cite{zhao2017pyramid} harvests different sub-region representations by concatenating the multi-level max-pooled features.
	%DeepLab~\cite{chen2017deeplab} proposes the atrous spatial pyramid pooling (ASPP), where multi atrous convolution layers with different rates capture multi-scale feature representations in parallel.

	\subsection{Efficient Neural Network Designing}
	
	Grouped convolution is first introduced in AlexNet~\cite{krizhevsky2012imagenet}
	% for the purpose of distributing the model over multi GPUs
	and then widely used in later networks, \textit{e.g.}, ResNeXt~\cite{xie2017aggregated}, for efficient computing.
	Depthwise convolution is the special case of grouped convolutions, where the feature channel of each group is single.
	Recent compact models running on mobile platforms such as MobileNetV2~\cite{sandler2018mobilenetv2} and ShuffleNet~\cite{zhang2018shufflenet}\cite{ma2018shufflenet} leverage the depthwise convolution extensively and achieve effective results.
	Particularly, ShuffleNet~\cite{zhang2018shufflenet} proposes a novel channel shuffling operation for fusing the features produced by different group of convolution filters.
	%This operation is more efficient and hardware-friendly then the ordinary $1\times1$ convolution.
	
	\section{Approach}
	
	In this section, we first propose that the optimal-size spatial-temporal kernel for video modeling can be decomposed into multi-scale spatial and temporal kernels of fixed-size fused by a learnable weight matrix.
	We further demonstrate that the weight matrix shall obey some explicit constraints and propose an optimized feature fusion layer.
	% under the judgment.
	Finally, we instantiate our method as an action perceptron synthesizer block and develop a new video modeling network upon the block with an efficient architecture and high performance.
	
	\subsection{Optimal Spatial-temporal Kernel Approximation}
	We first postulate there exists an \textit{ideal} video CNN, where each layer transforms a 4-dimensional input tensor $U$
	of size $C \times T \times W \times H$
	into an output tensor $V$ of the same size, $V = U * \hat{F}$,
	where $\hat{F}$ is the \textit{optimal} convolution kernel with the receptive field of size $\hat{T} \times \hat{W} \times \hat{H}$.
	The optimization goal of our method is to approximate $\hat{F}$ precisely under the assumption that 
	the upper bound of its receptive field is $L \times L \times L$.
	%However, speculating the optimal receptive field is challenging whereas 
	
	We propose to synthesize the \textit{unknow} $\hat{F}$ by approximating the produced $V$ as :
	\begin{align}\label{mk_ulti}
	\widetilde{V} = & \sum_{i=1}^{G} W^{'}(i) \odot ( \sum_{j=1}^{G} W(j) \odot U * F_s^{(2j-1) \times (2j-1) \times 1}) \notag \\ 
	&* F_t^{1 \times 1 \times (2i-1)} ,
	\end{align}
	where $G$ denotes the number of branches of different kernel, 
	$\odot$ denotes the channel-wise multiplication.
	% 	$W^{'}$ and $W$ are of size $G \times 1$ and $1 \times G$ respectively\footnote{We remove the dimension of size 1 in the next, \textit{i.e.}, $W^{'} = W^{'}[:,0], W = W[0,:]$.}.
	Each element of $W^{'}$ and $W$ is a $C$-length vector,
	indicating the channel-wise importance of the tensor produced by the kernels.
	% 	spatial and temporal kernels respectively.
	We show the derivation process in the \textit{supplementary material}.
	Although the representation seems to share some similarities with the R(2+1)D CNNs~\cite{tran2018closer},
	the focus of our method is to produce filters of \textit{diverse scales and shapes} instead of only saving the computation cost.

	The physical meaning behind the Eq. (\ref{mk_ulti}) is that the optimal-size spatial-temporal kernel $\hat{F}$ can be \textit{mimicked} by a bag of multi-scale spatial kernels $F_s=\{F_s^{1 \times 1 \times 1}, ..., F_s^{L \times L \times 1} \}$ followed by another bag of multi-scale temporal kernels
	$F_t = \{F_t^{1 \times 1 \times 1}, ..., F_t^{1 \times 1 \times L} \}$.
	The two bags are interacted by $W$ ($W^{'}$ is only related to the temporal kernels).
	
	In the next section, we discuss the constraints of $W$ under the group convolution setting,
	\textit{i.e.}, (1) $U = U_1 \oplus U_2 \oplus  ... \oplus U_G$, where $G$ denotes the group number.
	(2) Each kernel of $F_s$ and $F_t$ only performs on the one group.
	% 	of its input tensor.

	\subsection{Optimized Feature Fusion Layer}\label{sec_layer}

\begin{figure}
		%	\centering
		\begin{minipage}[htb]{1\linewidth}
			\centerline{\includegraphics[width=6cm]{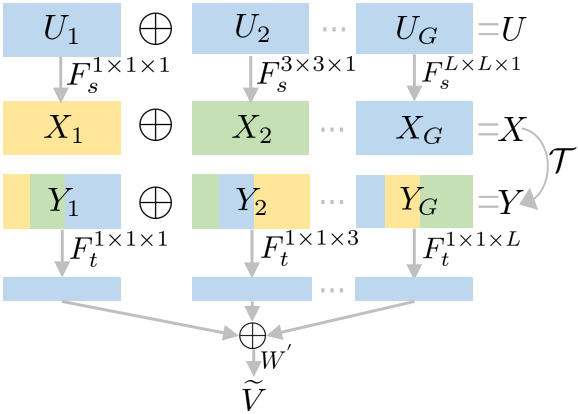}}
			\caption {
				Schematic representation of Eq. (\ref{mk_ulti}) under the group convolution setting.
				The core of the \textit{\textbf{optimized feature fusion layer}} is a constrained transformation matrix $\mathcal{T}$, which is \textit{reshaped} from the $W$ in Eq. (\ref{mk_ulti}).
				$\oplus$ denotes the channel concatenation operation.
			}
			\label{channel_inter_layer}
		\end{minipage}
		\vspace{-0.3cm}
	\end{figure}

	Given the feature $X = X_1 \oplus X_2 \oplus  ... \oplus X_G$ produced by spatial convolutions in Eq.~(\ref{mk_ulti}),
	$
	X_j = U_j * F_s^{(2j-1) \times (2j-1) \times 1},
	$
	where $j \in [1, G]$.
	%When performing a group of transformations $T = \{T_1, T_2, .., T_G\}$ on $X$, the output feature of each group is $Y_i = T_i(X_i)$ which captures only single scale receptive %field, where $T_i$ denotes the kernel of one scale.
	Now, we assume there exists a block transformation matrix
	$\mathcal{T} \in \mathbb{R}^{G\times G}$
	performing on $X$ to produce $Y = Y_1 \oplus Y_2 \oplus  ... \oplus Y_G $, as shown in Fig.~\ref{channel_inter_layer}.
	% 	$Y$ will be convolved with $G$ temporal convolution kernels $F_t$ of different sizes.
	Noting that $\mathcal{T}$ of size $G\times G$ is reshaped from $W$ in Eq. (\ref{mk_ulti}) of size $G$ by dividing the each element of it into $G$ groups.
	% 	The channel number of each group of $X$ and $Y$ are denoted as $c = \frac{C}{G}$.
	Each element of $\mathcal{T}$ is a $c$-element vector, which is consistent with the channel number of $X_j$ and $Y_i$, where $c = \frac{C}{G}$.
	%	 and $j \in  [1,G]$.
	We formulate the transformation process above as:
	$
	Y = \mathcal{T}\times X,
	$
	where the multiplication between each element of $\mathcal{T}$ and $X$ is channel-wise.
	In the following parts, we omit the domain of $i,j \in  [1,G]$.
	% 	We also assume that each group in $X$ and each weight vector in $\mathcal{T}$ are also sub-grouped into G sub-groups: 
	% 	$X_i = X_{i_1} \oplus ...\oplus X_{i_G}$
	% 	and 
	% 	$\mathcal{T}_{ij} = \mathcal{T}_{ij}(1)\oplus ... \oplus \mathcal{T}_{ij}(G)$
	% 	,
	% 	because some conventional feature transformation techniques, \textit{e.g.}, dropout~\cite{hinton2012improving} and channel shuffling~\cite{zhang2018shufflenet}, are performed on this sub-group level.
	%\begin{gather}\label{eq_block_mul}
	%\mathcal{T}_{ij} \odot X_i   = (\mathcal{T}_{ij}(1)\odot X_{i_1})\oplus ... \oplus (\mathcal{T}_{ij}(G)\odot X_{i_G}),
	%\end{gather}
	%Then, the multiplications between the weighting parameter $\mathcal{T}_{ij} \in \mathbb{R}^c$ and the feature $X_i$ in each group also follows the group decomposition calculation rules: 
	
	We quantify the routing paths between the spatial and temporal kernels of different size by:
	$
	ir-interactions = \sum_{i=1}^{G}\sum_{j=1}^{G}N_{X_i \to Y_j}, i \neq j,
	$
	where $N_{X_i \to Y_j}$ denotes the information flowed from ${X_i}$ to ${Y_j}$, which is measured by the number of feature channels.
	$ir-interactions$ indicates the number of the synthesized \textit{irregular-shaped} spatial-temporal kernels, compared to the regular-shaped kernels of size $3\times3\times3$ or $5\times5\times5$.
	In the experiment part, we show that this metric has strong impact on the performance. 
	
	\textbf{Optimization goal:}
	\noindent (1) \textbf{Spatial-temporal interaction richness.}
	We propose the interaction loss formulated as: 
	\begin{gather}\label{eq_loss1}
	\mathcal{L}_{interaction} = -\frac{1}{G^2} \sum_{i=1}^{G} \sum_{j=1}^{G} sigmoid(||\mathcal{T}_{ij}||_1),
	\end{gather}
	where $||\quad||_1$ denotes the $\ell1$ norm.
	This loss facilitates higher inter-group interaction richness of $X$ while regularizing the weights not exploding.
	By this way, the temporal convolutions after the fusion layer can receive very rich-scale spatial features.
	% taken input by the subsequent temporal convolutions .
	%The theoretical lower and upper bounds of $ir-interactions$ is $G^2$ and $G^2 \cdot c$ respectively.
	In practice, we find the $ir-interactions$ tends to be $G^2 \cdot c$ when utilizing this loss.
	
	\noindent (2) \textbf{High network capacity.}
	We propose the network capacity loss formulated as:
	\begin{gather}\label{eq_loss2}
	%	\mathcal{L}_{capacity} = \frac{1}{G^2} \sum_{i=1}^{G}\sum_{j=1}^{G}sigmoid(|\frac{Y_i \cdot Y_j}{||Y_i|| ||Y_j||}|),
	\mathcal{L}_{capacity} = \frac{1}{G^2} \sum_{i=1}^{G}\sum_{j=1}^{G}\frac{avg(Y_i) \cdot avg(Y_j)}{||avg(Y_i)||_2 ||avg(Y_j)||_2},
	\end{gather}
	where $avg$ denotes the spatial-temporal average pooling operation, $\cdot$ and $||\quad||_2$ denotes Dot product and $\ell2$ norm respectively.
	This loss facilitates the lower inter-group similarity within $Y$ and thus leads to the wider network of higher capacity in essence.
	
	\noindent \textbf{vs. Other feature fusion techniques.}
	We assume that each weight vector $\mathcal{T}_{ij}$ in $\mathcal{T}$ is also sub-grouped into G groups:
	$\mathcal{T}_{ij} = \mathcal{T}_{ij}(1)\oplus ... \oplus \mathcal{T}_{ij}(G)$.
	Then, under the framework of the proposed layer, channel shuffling~\cite{zhang2018shufflenet} can be represented as:
	\begin{gather}
	\{
	\begin{array}{lr}
	\mathcal{T}_{ij}(k) = \boldsymbol {1}, k=i, \\
	\mathcal{T}_{ij}(k) = \boldsymbol {0}, others.
	\end{array}
	,
	\end{gather}
	where $k \in [1,G]$,
	$\boldsymbol {1}$ and $\boldsymbol {0}$ denote \textit{all-one} and \textit{all-zero} vectors of size $\frac{c}{G}$ separately.
	Further comparing with other conventional feature fusion techniques as a special cases of our method is in the \textit{supplementary material}.

	\begin{figure}
		\centering
		\begin{minipage}[b]{1\linewidth}
			\centering
			\centerline{\includegraphics[width=7cm]{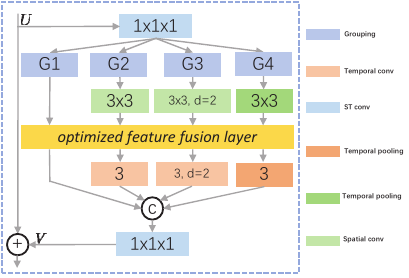}}
		\end{minipage}
		\caption {
			\small
			An example Action Perceptron Synthesizer of maximum receptive filed 5$\times$5$\times$5.
			{\textcopyright} and $\oplus$ denotes the channel-wise concatenation and element-wise summation operation respectively.
			All the filters are performed on the grouped features for efficient computation.
			The blocks marked with “d” are dilated convolutions.
			The \textit{optimzed feature fusion layer} can be implemented as a fully-connected layer regularized by $\mathcal{L}_{interaction}$ and $\mathcal{L}_{capacity}$.
		}
		\label{fig_aps}
	\end{figure}
	\subsection{Action Perceptron Synthesizer}\label{ASP}
	We wrap the approximated optimal spatial-temporal operation in Eq.~(\ref{mk_ulti}) into an action perceptron synthesizer that can be incorporated into many existing architectures.
	The synthesizer is defined as:
	$
	z_i = {V_i} + U_i,
	$
	where ${V_i}$ is given in Eq.~(\ref{mk_ulti}) and “+$U_i$” denotes a residual connection~\cite{he2016deep}.
	The residual connection allows us to insert the proposed synthesizer into any pre-trained model such as ResNet, without breaking its initial behavior (\textit{e.g.}, when ${W}$ in Eq.~(\ref{mk_ulti}) is initialized as zero).
	An example action perceptron synthesizer is illustrated in Fig.~\ref{fig_aps}.
	The \textit{optimized feature fusion layer} can be simply implemented as a fully connected layer while regularized by the two proposed losses.
	Only spatial convolutions are followed by the batch normalization (BN)~\cite{ioffe2015batch} and ReLU non-linearity~\cite{nair2010rectified}.
	We also add the max-pooling branch, which improves the performance obviously due to its complement for convolution operations.
	
	%The synthesizer is facilitated by the following several efficient CNN architecture designing practices.
	%(1) \textbf{Partial channel transformation}. 
	%Our method also only applies spatial-temporal modeling to a small proportion of the feature channels (1/4 or 1/8).
	%(2) \textbf{Group convolution}. Our method performs spatial or temporal modeling separately in each group for each kernel size, which further reduces the computation cost and the risk of over-fitting.
	%(3) \textbf{Dilated convolution}. We also replace the large-size convolution kernels with the dilated counterparts of them to further reduce the parameter number.

	%by leveraging the  to enhance the RF-L-Inception structure with asymmetric spatial-temporal modeling ability.
	% above and several other light-weight CNN designing practices, as shown in Fig. \ref{fig_ast_blk}.
	%To alleviate the cost of the vanilla 3D-MS-Large structure while preserving i
	
	\noindent\textbf{Video Perceptron Synthesis Networks.}
	The proposed action perceptron synthesizer is flexible and can be easily integrated with most of the current 2D or 3D CNNs.
	% stacked by 2D or 3D convolutions.
	More specifically, we adopt 2D-ResNet-50~\cite{he2016deep} as the backbone networks and insert the proposed synthesizer between the residual blocks.
	%We customize the spatial-temporal shape diversity by leveraging a different number of this block.
	The final prediction is a simple average pooling of the results from each frame.
	We conduct extensive experiments on the different variants of it.
	%We also conduct experiments to show if other more sophisticated late feature fusion method such as TRN~\cite{zhou2018temporal} and ECO~\cite{zolfaghari2018eco} can improve the performance further.
	
	\noindent\textbf{End-to-end learning for action recognition}.
	Finally, we apply the proposed networks to the action recognition task.
	% 	by optimizing the classification objective and the \textit{optimized feature fusion layer} simultaneously.
	The total loss is given by 
	$
	\mathcal{L} = \mathcal{L}_{classification} + \alpha\mathcal{L}_{interaction} + \beta\mathcal{L}_{capacity},
	$
	where $\mathcal{L}_{classification}$ is the cross-entropy loss, $\alpha$ and $\beta$ are the balancing weights.
	
	\section{Experiments}
	\subsection{Video Datasets}
	
	\noindent \textbf{Something-Something.}
	This dataset includes v1~\cite{goyal2017something} and v2~\cite{mahdisoltani2018fine}.
	%The required spatial-temporal receptive-field varies hugely across different fine-grained level actions as shown in Fig.~\ref{fig_1} (a),
	%which is very suitable for verifying the flexible spatial-temporal modeling ability of the proposed method.
	We mainly conduct ablation experiments and justify each component on Something-Something v1 dataset. 
	
	\noindent \textbf{Kinetics.}
	Kinetics~\cite{carreira2017quo} is a challenging human action recognition dataset.
	%Compared to the actions in Something-Something, 
	The actions in this dataset mainly rely on the appearance of the objects and the background scenes to be discriminated.
	%We conduct the experiments on Kinetics-400~\cite{carreira2017quo} because there are many well known baseline methods benchmarked on this dataset.
	
	\noindent \textbf{Diving48.}
	Diving48~\cite{li2018resound} is a new dataset with more than 18K video clips for 48 unambiguous diving classes, requiring multi-scale temporal modeling.
	%Therefore, we conduct experiments on this dataset to verify the multi-scale spatial-temporal modeling ability of our method comprehensively.
	We report the accuracy on the official train/val split. 
	
	\subsection{Implementation Detail}
	
	We implement our model in Pytorch~\cite{paszke2019pytorch}. We adopt ResNet50~\cite{he2016deep} pretrained on ImageNet~\cite{deng2009imagenet} as the backbone.
	The parameters within the action perceptron synthesizers are randomly initialized.
	The synthesizers are inserted after the $conv_{1-4}$ if no specified otherwise.
	For the temporal dimension of the input clips, we use the sparse sampling method described in TSN~\cite{wang2016temporal}.
	%Specifically, the videos are first divided into several segments of equal duration, and then one snippet is randomly sampled from its corresponding segment.
	%The snippet forms the clip input to the networks.
	For spatial dimension, the short-side of the input frames are resized to $256$ and then cropped to $224\times224$. We do random cropping and flipping as data augmentation during the training.
	%It's worth to note that we do not perform horizontal flipping on some moving direction related action classes such as “moving something from left to right”.
	We train the network with a batch-size of 64 on 8 NVIDIA GTX-2080Ti GPUs and optimize it using SGD with an initial learning rate of 0.01 for 50 epochs and decay it by a factor of 10 every 10 epochs. The total training epochs are about 80.
	To train the total
	loss, we set the balancing weights of the losses as: $\alpha = 0.01$ and $\beta = 0.001$ by grid searching.
	The dropout ratio is set to be 0.3 as in~\cite{luo2019grouped}. During the inference, we sample the middle frame in each segment and do center crop for each frame. We report the results of \textbf{1 crop} unless specified.
	Note that many state-of-the-art methods report their final performances with 5 or 10 crops, which enlarge the inference-time computation cost by 5 or 10 times.
	Moreover, we only use \textbf{RGB modality} as the input to our model.
	%unlike two-stream networks~\cite{simonyan2014two}\cite{feichtenhofer2016convolutional} which use both RGB and optical-flow modalities.

	\begin{figure}[t]
		\centering
		\begin{minipage}[b]{0.8\linewidth}
			\centering
			\centerline{\includegraphics[width=8cm]{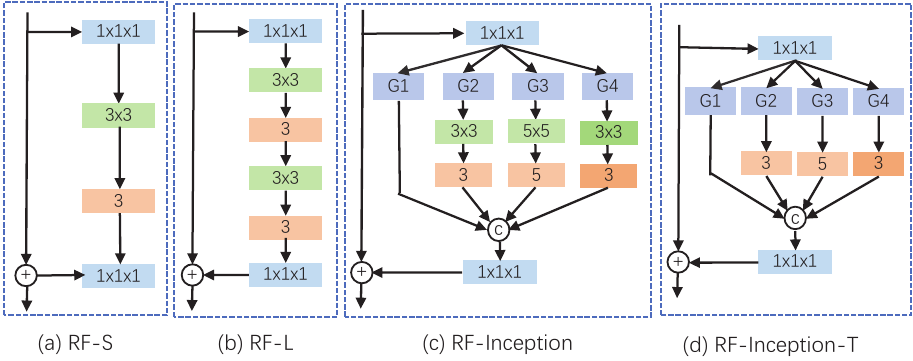}}
		\end{minipage}
		\caption {
			\small
			Illustrations of the baseline methods.
			The representation signs are the same meaning with Fig.~\ref{fig_aps}.
			“RF” is abbreviated from “receptive field”.
			The convolutions of kernel size 5 are also implemented as their dilated counterparts.
		}
		\vspace{-0.3cm}
		\label{fig_base_arch}
	\end{figure}

	\setlength{\tabcolsep}{5pt}
	\renewcommand{\arraystretch}{0.8}
	\begin{table*}[htbp]
		\centering
		\small
		\begin{tabular}{|c|c|c|c|c|c|c|c|c|}
			\hline
			\multirow{2}{*}{\tabincell{c}{Methods} } & 
			\multirow{2}{*}{\tabincell{c}{Backbone} } & 
			\multirow{2}{*}{\tabincell{c}{Maximum \\RFS} }&
			\multirow{2}{*}{\tabincell{c}{Synthesized \\kernel shapes} } &
			\multirow{2}{*}{\tabincell{c}{Irregular \\shape?} } &
			\multirow{2}{*}{\tabincell{c}{Multi-scale?} } & 
			\multirow{2}{*}{\tabincell{c}{Parameters} } & 
			\multicolumn{2}{c|}{Sth v1}  \\
			\cline{8-9} 	&&&&&&& \scriptsize{Top1 (\%)} & \scriptsize{Top5 (\%)} \\ 
			\hline\hline
			TSN~\cite{wang2016temporal}$^\dag$ &ResNet50&-&-&-&-& 23.87M & 14.90 & 36.97  \\
			TSM~\cite{lin2019tsm}$^\dag$ &ResNet50&$1\times3$&1&\checkmark&-&23.87M& 42.11 & 71.01 \\
			GST~\cite{luo2019grouped}$^\dag$ &ResNet50&$3\times3$&1&-&-& 21.04M&  42.36 & 71.42 \\
			GST~\cite{luo2019grouped}$^\dag$ &ResNet101&$3\times3$&1&-&-& 37.52M&  41.16 & 69.76 \\
			\hline
			RF-S &ResNet50&$3\times3$&1&-&-&27.57M    & 43.72 & 73.14 \\
			RF-L &ResNet50&$5\times5$&1&-&-&27.35M   & 43.67 & 71.88 \\
			RF-L-Inception &ResNet50&$5\times5$&4&-&\checkmark &27.56M & 43.84 & 72.59 \\
			RF-L-Inception-T &ResNet50&$1\times5$&4&\checkmark&\checkmark &27.28M & 44.23 & 72.87 \\
			\textbf{Ours} &ResNet50&\textbf{$5\times5$}&16& \checkmark &\checkmark &27.55M& \textbf{45.44} & \textbf{74.12}   \\
			%\textbf{Ours} &ResNet101&\textbf{$5\times5$}& \checkmark &\checkmark &-& \textbf{-} & \textbf{-}   \\
			\textbf{Ours-1/16} &ResNet50&\textbf{$5\times5$}&16& \checkmark &\checkmark &24.05M& \textbf{44.00} & \textbf{72.65}   \\
			\hline
		\end{tabular}
		\caption{
			\small
			Results of inserting different spatial-temporal blocks to 2D ResNet-50.
			%		“RFS” is abbreviated from receptive field size.
			%		All of our baselines outperform the previous works significantly.
			The evolution of the performance improvement can be explained by diverse shape synthesized kernels, especially the introduction of irregular shape kernels, and multi-scale designing. 
			$^\dag$ indicates that the results are reproduced under the same input size with us.
		}
		
		\label{tab_st_model}
	\end{table*}

	\setlength{\tabcolsep}{2pt}
	\renewcommand{\arraystretch}{0.8}
	\begin{table*}[tbp]
		\begin{minipage}{0.6\textwidth}
			
			\centering
			\small
			\begin{tabular}{|p{3cm}<{\centering}|c|c|c|c|c|}
				\hline
				\multirow{2}{*}{\tabincell{c}{ Methods } } &
				\multirow{2}{*}{\tabincell{c}{$ir-$ \\ $interactions$ } } & 
				\multirow{2}{*}{\tabincell{c}{Dropout \\ rate } } & 
				\multirow{2}{*}{\tabincell{c}{Parameters  } } & 
				\multicolumn{2}{c|}{Sth v1}  \\
				\cline{5-6} 
				&&&& \scriptsize{Top1 (\%)} & \scriptsize{Top5 (\%)} \\ 
				\hline\hline
				Channel grouping & 0 & 0& 0M & 43.84 & 72.59 \\
				\hline
				Channel dropout & $1856$ & $0.75$ &0M& 44.64 & 73.03 \\
				Channel shuffling & $1856$ &\textbf{0}&\textbf{0M} & 45.01 & 73.53\\
				Ours & $<1856$ &$>0$ &0.086 M& \textbf{45.44} & \textbf{74.12} \\
				%		Ours+softmax & $<1856$ &$>0$ &0.086 M& - & - \\
				Ours+$\mathcal{L}_{i}$ & $\sim 1856$ &$>0$ &0.086 M& 46.12 & 74.92\\
				Ours+ $\mathcal{L}_{c}$ & $<1856$ &$0$ &0.086 M& 45.88 & 74.22 \\
				Ours+\textbf{$\mathcal{L}_{i}$ and $\mathcal{L}_{c}$ }& $\sim 1856$ &$0$ &0.086 M& \textbf{46.41} & \textbf{75.01}   \\
				\hline
			\end{tabular}
			\caption{
				\small
				Comparison of different feature fusion methods.
				Larger $ir-interactions$ and lower dropout rate bring more performance gain.
				$\mathcal{L}_{i}$ and $\mathcal{L}_{c}$ denotes $\mathcal{L}_{interaction}$ and $\mathcal{L}_{capacity}$ respectively.
				Architecture details are in \textit{supplementary material}.
			}
			\label{tab_ft_inter}
		\end{minipage}
		\begin{minipage}{0.38\textwidth}
			\centering
			\small
			\begin{tabular}{|l|c|c|c|}
				\hline
				\multirow{2}{*}{\tabincell{c}{Design } } & \multirow{2}{*}{\tabincell{c}{Parameter \\ number } } 
				& \multicolumn{2}{c|}{Sth v1}  \\
				\cline{3-4} 	&& \scriptsize{Top1 (\%)} & \scriptsize{Top5 (\%)} \\ 
				\hline
				Avg Pool &27.55M& 45.42 & 73.98   \\
				Max Pool &27.55M& \textbf{45.44} & \textbf{74.12}   \\
				\hline
				Without Max Pool &27.55M& 44.19 & 73.45   \\
				Max Pool &27.55M& \textbf{45.44} & \textbf{74.12}   \\
				\hline
				Without inter ReLU &27.55M& 44.65 & 73.30   \\
				Inter ReLU &27.55M& \textbf{45.44} & \textbf{74.12}   \\
				\hline
				Without dilation &27.95M& \textbf{45.54} & 73.09   \\
				Dilation &\textbf{27.55M}& 45.44 & \textbf{74.12}   \\
				\hline
				(1+1+1)D &\textbf{27.45M}& 45.17 & 73.77   \\
				(2+1)D &27.55M& \textbf{45.44} & \textbf{74.12}   \\
				\hline
			\end{tabular}
			\caption{
				\small
				Detailed design of the proposed block.
			}
			\label{tab_arch_details}

		\end{minipage}
	\end{table*}
	
	\subsection{Ablation Study}
	We conduct extensive ablation studies on the Something-Something V1~\cite{goyal2017something} dataset to demonstrate the effectiveness of every aspects of our method.
	All the synthesizers in these experiments are of maximum receptive field $5\times5\times5$ as shown in Fig.~\ref{fig_aps} and performed on the $1/4$ proportion of input features if no specified otherwise.
	To facilitate the training process, in these experiments, we adopt a smaller input resolution: the short-side of the input frames are resized to $128$ and then the frames are cropped to $112\times112$.
	
	%The batch size is of 120.
	%For the convenience of expression, the five blocks of ResNet-50 are denoted as conv1, conv2, conv3, conv4 and conv5 respectively.
	To emphasize the importance of two important philosophies in our method: (a) \textit{multi-scale modeling} and (b) \textit{optimal-size kernels synthesizing} especially the \textit{irregular-shaped} kernels, we compare against the following baselines:
	(1) The vanilla TSN~\cite{wang2016temporal},
	(2) \textbf{RF-S model} shown in Fig.~\ref{fig_base_arch} (a),
	(3) \textbf{RF-L model} shown in Fig.~\ref{fig_base_arch} (b),
	(4) \textbf{RF-L-Inception model} shown in Fig.~\ref{fig_base_arch} (c),
	(5) \textbf{RF-L-Inception-T model} shown in Fig.~\ref{fig_base_arch} (d).
	The architecture details are in the \textit{supplementary material}.
	%To keep the parameter number of the baselines consistent with our method, we set the \textit{feature proportion} (defined in Sec.~\ref{ASP}) as $1/6$, $1/8$, $1/4$ and $1/4$ for the RF-S, RF-L, RF-L-Inception and RF-L-Inception-T models respectively.

	\subsubsection{\textbf{The superiority of action perceptron synthesizer}}
	
	As shown in Tab.~\ref{tab_st_model}, our method mine the spatial-temporal reasoning information much better.
	Compared with the RF-L-Inception model, our method improves the performance significantly with almost no extra computation cost,
	thanks to the kernels of much more diverse shapes (16 vs. 4), especially the irregular shape, introduced by the \textit{synthesizers}.
	
	Also, \textit{multi-scale} modeling widely adopted in the modern CNNs only contributes marginal performance gain: \textit{0.17\%} (RF-L vs. RF-L-Inception)
	whereas \textit{rich kernel shape} brings conspicuous improvement: \textbf{\textit{1.60\%}} (RF-L-Inception vs. our method) in terms of Top1 accuracy.
	Interestingly, only enlarging the maximum receptive field in a naive way, \textit{i.e.}, from RF-S model to RF-L model, the performances are degraded slightly.
	We conjecture that the degradation is caused by the smaller enhanced spatial-temporal \textit{feature proportion} of RF-L model ($1/6$ vs. $1/8$), which will be discussed in \textit{supplementary material}.
	% the RF-S model which only performs spatial-temporal modeling on the $\frac{1}{8}$ features output from/ 
	%Moreover, all of our baselines outperform TSM network obviously, which exchanges part of the channels along temporal dimension and invalidates this part of the features, indicating the importance of keeping the original feature hierarchy of the 2D backbone CNNs.
	%To eliminate the influence of the extra spatial filters introduced by our method, we propose the  baseline consists of only temporal convolutions.
	The large performance improvement of RF-L-Inception-T upon TSM also proves the merits of \textit{learnable multi-scale} temporal modeling.
	To compare with TSN and TSM rigorously, we also set the \textit{feature proportion} of our method as $1/16$, resulting the comparable parameter number as them.
	Our method outperforms TSM by \textbf{1.89\%} in terms of Top1 accuracy although TSM performs temporal modeling on more feature maps (1/8 of the input feature maps).
	Compared to GST, our method enhances the spatial features with the spatial-temporal information by \textit{adding} instead of \textit{replacing}.  
	%Partially similar to our method, GST converts some 2D spatial features to 3D spatial-temporal features 
	% while utilizing 3D convolutions of $3\times3\times3$ receptive field size.
	All of our baselines even including the RF-S model outperform GST, proving the superiority of retaining the complete spatial appearance prior learned on image tasks.
	GST has less parameters even than the simplest TSN because it sacrifices the channel number of spatial features,
	showing over-fitting when leveraging the ResNet101 as the backbone network.
	
	% 	which may lead to serious performance degradation on very large
	% 	scale video datasets such as Kinetics.

	%GST has less trainable parameters than our method under the same backbone network because they groups the original spatial features within the bottleneck.
	%Thus, we compare our model based on ResNet50 with their model based on ResNet101.
	%As shown in Tab.~\ref{tab_st_model}, TODOTODO

	%%%%interactino of the above table are computed by (8+32+64+128)
	%%%% paramter = (8^2+32^2+64^2+128^2)
	\subsubsection{\textbf{The superiority of optimized feature fusion layer}}
	As shown in Tab.~\ref{tab_ft_inter}, when utilizing plain channel grouping operation between the spatial and the temporal filters, our method degenerates to the RF-L-Inception baseline and shows the most inferior performance for not synthesizing rich-scale spatial-temporal kernels.
	The synthesizing process relies on the irregular routing paths quantified by $ir-interactions$, which is \textit{0} in this situation.
	The performance improves consistently with larger $ir-interactions$ and lower dropout rate.
	Notably, the dropout based fusion method (detailed in \textit{supplementary material}) still outperforms the baseline largely while discarding a large proportion of spatial features,
	implying the synthesized \textit{irregular-shaped} kernels measured by $ir-interactions$ have more direct impact on the final performance.
	Channel shuffling demonstrates the best performance under the premise of not introducing extra parameters.
	However, its performance is worse than our method even without any regularization ($45.01\%$ vs. \textbf{45.44\%}) because it is un-learnable and can not adjust the weights of the routing path dynamically for synthesizing the optimal kernels in a data-driven way.
	With the both two proposed regularizations ($\mathcal{L}_{interaction}$ and $\mathcal{L}_{capacity}$) activated,
	our method improves the performance by near another \textbf{1\%}.
	% 	our method depicts the best performance.

	\subsubsection{\textbf{Studies on network details}}
	In the \textit{supplementary material}, we show the ablation studies on \textit{Where to insert the block?} and \textit{What proportion of features need to be enhanced with space-temporal features?}
	Then, we demonstrate the studies on the detailed design of our block in Tab.~\ref{tab_arch_details},
	the pooling operation improves the performance significantly (\textbf{1.25\%} Top1 accuracy) while our method is not sensitive to the specific implementation of the operation.
	%Either average pooling or max pooling can achieve excellent performance.
	%Max pooling has slight advantage over average pooling because the regions of focusing objects and the key frames related to action state changes are only a small proportion of the input video data.
	The extra non-linearity introduced by the intermediate ReLU operations between spatial- and temporal- filters also benefits the performance obviously, which is consistent with the conclusion from previous work~\cite{tran2018closer}.
	The dilated convolution outperforms the ordinary convolution by \textbf{1.24\%} in terms of Top5 accuracy even with less parameters, showing better generalization ability.
	Finally, we also try to decompose the spatial convolutions.
	As shown in the last block of Tab.~\ref{tab_arch_details},
	compared to the (2+1) D setting, the (1+1+1) D truly reduces negligible number of parameters but also degrades the performance.
	
	\subsection{Comparison with State-of-the-Art}

	\begin{table*}[ht]
		%		\begin{minipage}{0.7\textwidth}
		\renewcommand{\arraystretch}{0.8}
		\centering
		\small
		\begin{tabular}{|c|c|c|c|c|c|c|}
			\hline
			\textbf{Method} & \textbf{Backbone} & \textbf{Pre-train} & \textbf{\#Frames} & \textbf{GFLOPs} & \textbf{Top1 (\%)} & \textbf{Top5 (\%)} \\
			\hline \hline
			
			I3D~\cite{carreira2017quo}  & 3D-ResNet-50 & Kinetics & 32$\times$3$\times$2 & 153$\times$3$\times$2 & 41.6 &72.2 \\ \hline
			%		Non-local~\cite{wang2018non}  & 3D-ResNet-50 & Kinetics & 32$\times$3$\times$2 & 168$\times$3$\times$2 & 44.4&76.0 \\ \hline
			GCN+Non-local~\cite{wang2018videos} & 3D-ResNet-50 & Kinetics & 32$\times$3$\times$2 & 303$\times$3$\times$2 & 46.1&76.8 \\ \hline
			ECO(En)~\cite{zolfaghari2018eco} & BNInc + 3D-ResNet-18 & Kinetics & 92$\times$1$\times$1 & 267$\times$1$\times$1 & 46.4 &- \\ \hline
			%		ECO(En)+flow~\cite{zolfaghari2018eco} & BNInc + 3D-ResNet-18 & Kinetics & 92+92 & N/A & 49.5 &- \\ \hline
			\hline \hline	
			TSN~\cite{wang2016temporal}  & BN-Inception & ImageNet & 8$\times$1$\times$1 & 16$\times$1$\times$1 & 19.5 &- \\ 
			%		TSN~\cite{wang2016temporal}  & BN-Inception & ImageNet & 16$\times$1$\times$1 & 32.73$\times$1$\times$1 & 17.5 &- \\ \hline
			MultiScale TRN~\cite{zhou2018temporal}  & BN-Inception & ImageNet & 8$\times$1$\times$1 & 16.37$\times$1$\times$1 & 34.4 &63.2  \\ 
			%		MultiScale TRN+flow~\cite{zhou2018temporal}  & BN-Inception & ImageNet & 8$\times$1$\times$1 & N/A & 42 &-  \\ \hline
			R(2+1)D~\cite{sudhakaran2020gate}  & ResNet-34 & Sports1M & 32$\times$1$\times$1 & 152$\times$1$\times$1 & 45.7 &- \\ \hline
			S3D-G~\cite{xie2018rethinking}  & InceptionV1 & ImageNet & 64$\times$1$\times$1 & 71.38$\times$1$\times$1 & 48.2&78.7 \\ \hline
			%			STM~\cite{jiang2019stm}   & ResNet-50 & ImageNet & 8$\times$3$\times$10 & 33$\times$3$\times$10 & 49.2&79.3  \\ 
			STM~\cite{jiang2019stm}   & ResNet-50 & ImageNet & 16$\times$3$\times$10 & 67$\times$3$\times$10 & 50.7&80.4  \\ \hline
			TSM~\cite{lin2019tsm}   & ResNet-50 & Kinetics & 8$\times$1$\times$1 & 33$\times$1$\times$1 & 45.6 &74.2 \\ 
			TSM~\cite{lin2019tsm}   & ResNet-50 & Kinetics & 16$\times$1$\times$1 & 65$\times$1$\times$1 & 47.2 &77.1  \\ \hline
			GST~\cite{luo2019grouped}  & ResNet-50 & ImageNet & 8$\times$1$\times$1 & 29.5$\times$1$\times$1 & 47.0&76.1 \\
			GST~\cite{luo2019grouped}  & ResNet-50 & ImageNet & 16$\times$1$\times$1 & 59$\times$1$\times$1 & 48.6&77.9 \\ \hline
			TEA~\cite{li2020tea}  & ResNet-50 & ImageNet & 8$\times$1$\times$1 & 35$\times$1$\times$1 & 48.9&78.1 \\
			TEA~\cite{li2020tea}  & ResNet-50 & ImageNet & 16$\times$1$\times$1 & 70$\times$1$\times$1 & 51.9&80.3 \\ 
			TEA~\cite{li2020tea}  & ResNet-50 & ImageNet & 16$\times$3$\times$10 & 70$\times$3$\times$10 & 52.3&81.9 \\  \hline
			%			GSM~\cite{sudhakaran2020gate}  & InceptionV3 & ImageNet & 8$\times$1$\times$1 & 26.85$\times$1$\times$1 & 49.0&- \\
			%			GSM~\cite{sudhakaran2020gate}  & InceptionV3 & ImageNet & 16$\times$1$\times$1 & 53.7$\times$1$\times$1 & 50.6&- \\ 
			%			GSM~\cite{sudhakaran2020gate}  & InceptionV3 & ImageNet & 8$\times$2+12$\times$2+16+24 & 268.47 & \textbf{55.1}&- \\ 
			\hline \hline
			\multirow{4}{*}{Ours} 
			& ResNet-50 & ImageNet & 8$\times$1$\times$1$^\dagger$ & \textbf{9.12$\times$1$\times$1} & 47.2&75.6 \\\cline{2-7}
			& ResNet-50 & ImageNet & 8$\times$1$\times$1 & 36.19$\times$1$\times$1 &\textbf{50.8} &\textbf{80.6} \\ \cline{2-7}
			& ResNet-50 & ImageNet & 16$\times$1$\times$1 & 72.38$\times$1$\times$1 & \textbf{53.6} & \textbf{83.1} \\ \cline{2-7}
			& ResNet-50 & ImageNet & 16$\times$3$\times$10 & 36.19$\times$3$\times$10 & \textbf{54.1} & \textbf{84.3} \\ \cline{2-7}
			\hline
		\end{tabular}
		\caption{
			%			\tiny
			Comparison to state-of-the-art on Something-V1 validation set.
			$^\dagger$ indicates the spatial resolution is of 112$\times$112.
			%			$-$ indicates the paper didn't provide the results.
		}
		\label{tab_something_sota}
	\end{table*}
	
	\subsubsection{Something-V1}

	The recognition performance obtained by our method is compared with state-of-the-art approaches that just use RGB frames, as shown in Tab.~\ref{tab_something_sota}.
	The maximum RFS is 5$\times$5$\times$7 by grid searching.
	The results \textit{w.r.t.} the other RFSs is in the \textit{supplementary material}.
	We also adopt this hyper-parameter for the other two datasets.
	The first block of the table shows the approaches that utilize Full-3D CNNs.
	The second block of the table lists methods leveraging 2D CNN or efficient 3D CNN implementation.
	From the table, it can be seen that our method results in an absolute gain of \textbf{+34.1\%} (19.5\% vs. 53.6\%) over the TSN baseline.
	Our method performs better than 3D CNNs or heavier backbones with considerably less number of FLOPs.
	% although those approaches use external data for pre-training or ensemble the results from optical flow input.
	Under several common testing protocols,
	% 	\textit{i.e.}, temporal length of 8/16 with single/multi crop, 
	our method outperforms the most recent works significantly with comparable computational budget.
	%	 \textit{e.g.}, GST, GSM and TEA.
	%	The ensemble model of GSM outperforms our method largely with acceptable computational cost.
	%	However, it is not practical for initializing 4 models simultaneously and consuming large GPU memory.
	Moreover, when adopting inputs of lower spatial resolution, \textit{i.e.}, $112\times112$, our method still achieves competitive result (\textbf{47.2\%} top1 accuracy) with the recent methods such as TSM and GST, with much lower computation cost (\textbf{$<$ 30\%} of them).
	The analysis on the sensitivity of the methods to the input spatial resolution are in the \textit{supplementary material} and demonstrates that our method show the lowest sensitivity.

	\subsubsection{Kinetics-400}
	As shown in Tab.~\ref{tab_k400_sota}, our method also captures rich object appearance cues effectively.
	Our method achieves inferior performance compared to some 3D CNN methods, \textit{i.e.}, I3D and SlowFast networks.
	However, they both adopt much deeper and heavier 3D-ResNet-101 as the backbone network.
	They also leverage the inefficient non-local operation to model the long-range temporal dependencies.
	When comparing with the methods based on 2D CNNs, our method outperforms them largely and demonstrates the best trade-off between the action recognition accuracy and the computation cost.
	
	\begin{table*}[htbp]
		
		\renewcommand{\arraystretch}{0.8}
		\centering
		\small
		\begin{tabular}{|c|c|c|c|c|c|c|}
			\hline
			\textbf{Method} & \textbf{Backbone} & \textbf{Pre-train} & \textbf{\#Frames} & \textbf{GFLOPs} & \textbf{Top1 (\%)} & \textbf{Top5 (\%)} \\
			\hline \hline
			
			I3D~\cite{carreira2017quo}  & Inception V1 & ImageNet & 64$\times$N/A$\times$N/A & 108$\times$N/A$\times$N/A & 72.1 &90.3 \\ 
			%		I3D~\cite{carreira2017quo}  & Inception V1 & None & 64$\times$N/A$\times$N/A & 108$\times$N/A$\times$N/A & 67.5 &87.2 \\ 
			I3D+NL~\cite{wang2018non}  & 3D-ResNet-101 & ImageNet & 32$\times$6$\times$10 & 359$\times$6$\times$10 &\textbf{ 77.7}&93.3 \\ \hline
			ECO(En)~\cite{zolfaghari2018eco} & BNInc + 3D-ResNet-18 & None & 92$\times$1$\times$1 & 267$\times$1$\times$1 & 70.0 &- \\ \hline
			%		SlowOnly~\cite{feichtenhofer2019slowfast} & 3D-ResNet-50 & None & 8$\times$3$\times$10 & 41.9$\times$3$\times$10 & 74.8 & 91.6 \\
			SlowFast~\cite{feichtenhofer2019slowfast} & 3D-ResNet-50 & None & (4+32)$\times$3$\times$10 & 36.1$\times$3$\times$10 & 75.6 & 92.1 \\ 
			SlowFast+NL~\cite{feichtenhofer2019slowfast} & 3D-ResNet-101 & None & (8+16)$\times$3$\times$10 & 234$\times$3$\times$10 & \textbf{79.8} & \textbf{93.9} \\ \hline
			
			\hline \hline	
			TSN~\cite{wang2016temporal}  & BN-Inception & ImageNet & 25$\times$10$\times$1 & 53$\times$10$\times$1 & 69.1 &88.7 \\ \hline
			%		TSN~\cite{wang2016temporal}  & Inception v3 & ImageNet & 25$\times$10$\times$1 & 80$\times$10$\times$1 & 72.5 &90.2 \\ \hline
			R(2+1)D~\cite{tran2018closer}  & ResNet-34 & None & 32$\times$1$\times$10 & 152$\times$1$\times$10 & 72.0 & 90.0 \\ \hline
			TSM~\cite{lin2019tsm}   & ResNet-50 & ImageNet & 8$\times$3$\times$10 & 33$\times$3$\times$10 & 74.1 &- \\ 
			TSM~\cite{lin2019tsm}   & ResNet-50 & ImageNet & 16$\times$3$\times$10 & 65$\times$3$\times$10 & 74.7 &-  \\ \hline
			STM~\cite{jiang2019stm}   & ResNet-50 & ImageNet & 16$\times$3$\times$10 & 67$\times$3$\times$10 & 73.7&91.6  \\ \hline
			
			TEA~\cite{li2020tea}  & ResNet-50 & ImageNet & 8$\times$3$\times$10 & 35$\times$3$\times$10 & 75.0&91.8 \\
			TEA~\cite{li2020tea}  & ResNet-50 & ImageNet & 16$\times$3$\times$10 & 70$\times$3$\times$10 & 76.1&92.5 \\  \hline
			\hline \hline
			\multirow{4}{*}{Ours} 
			& ResNet-50 & ImageNet & 8$\times$1$\times$1 & 36.19$\times$1$\times$1 & 73.1 & 90.9 \\ \cline{2-7}
			& ResNet-50 & ImageNet & 8$\times$3$\times$10 &  36.19$\times$3$\times$10 & \textbf{76.2} & \textbf{92.8} \\ \cline{2-7}
			& ResNet-50 & ImageNet & 16$\times$1$\times$1 & 72.38$\times$1$\times$1 & 75.2 & 92.3 \\ \cline{2-7}
			& ResNet-50 & ImageNet & 16$\times$3$\times$10 &  72.38$\times$3$\times$10 & \textbf{77.5} & \textbf{93.1} \\ \cline{2-7}
			\hline
		\end{tabular}
		\caption{Comparison to state-of-the-art on Kinetics-400.
			%			$-$ indicates the paper didn't provide the results.
		}
		\label{tab_k400_sota}
	\end{table*}

	\subsubsection{Diving48}
	\begin{table}[htb]
		\renewcommand{\arraystretch}{0.8}
		\centering
		\small
		\begin{tabular}{|c|H|c|c|}
			\hline
			\textbf{Method} & \textbf{Pre-training} & \textbf{\#Frames} & \textbf{Top1 (\%)} \\ \hline \hline
			TSN (from \cite{li2018resound}) & ImageNet & 8 & 16.77 \\ \hline
			TRN (from \cite{li2018resound})& ImageNet & 8 & 22.8 \\ \hline
			%			C3D (from \cite{li2018resound})& ImageNet & 64 & 27.6 \\ \hline
			%		R(2+1)D (from \cite{bertasius2018learning}) & Kinetics&- & 28.9 \\ \hline
			%		DiMoFs~\cite{bertasius2018learning} & Kinetics+PoseTrack&- & 31.4 \\ \hline
			P3D-ResNet50 (from~\cite{luo2019grouped}) & ImageNet&16 & 32.4 \\ \hline
			C3D-ResNet50 (from~\cite{luo2019grouped}) & ImageNet &16& 34.5 \\ \hline
			~\cite{kanojia2019attentive} & ImageNet &64& 35.64 \\ \hline
			GST~\cite{luo2019grouped} & ImageNet &16& 38.8  \\ \hline 
			CorrNet-101~\cite{wang2020video} & - &32$\times$10& 38.6 \\ \hline
			GSM~\cite{sudhakaran2020gate} & ImageNet &16$\times$2& 40.27 \\ \hline \hline 
			Ours & ImageNet &16& \textbf{39.84} \\ \hline
			Ours & ImageNet &16$\times$2&  \textbf{41.02} \\ \hline
		\end{tabular}
		\caption{Comparison to state-of-the-art on Diving48.}
		\label{tab_diving48_sota}
		\vspace{-0.3cm}
	\end{table}
	
	To prove that our method can model long-term complex fine-grained motion cues and is not prone to over-fitting on few training samples, we test out method on Diving48.
	We input 16 frames to the network and sample only \textit{one} or \textit{two} clip from the video during inference.
	The results are shown in Tab.~\ref{tab_diving48_sota}.
	Our method achieves significant improvement over the most recent state-of-the-arts, \textit{i.e.}, over \textbf{1\%} under the input of 16 frames.
	%Our method achieves a recognition accuracy of \textbf{41.2\%}, an improvement of \textbf{+1.3\%} over previous  [27].
	%It's also worth to note that GSM leverages 2 clips as input but achieves inferior results.

	\subsection{Visualization}
	
	\begin{figure}[htbp]
		\centering
		\begin{minipage}[b]{0.8\linewidth}
			\centering
			\centerline{\includegraphics[width=8.6cm]{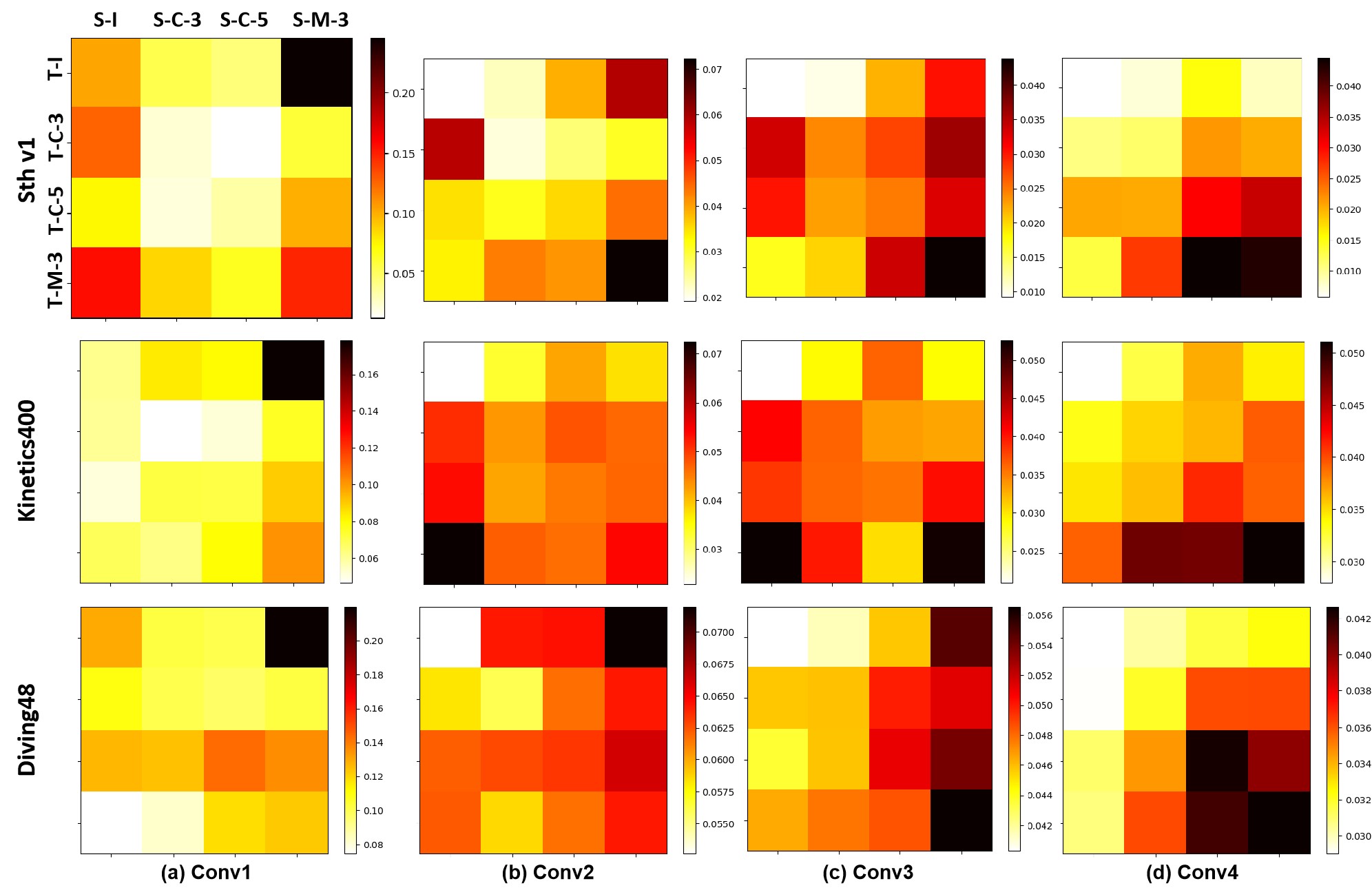}}
		\end{minipage}
		\caption {
			\small
			Visualization of spatial-temporal filter interactions of different layers (\textit{i.e.}, Conv$1\sim4$) over multiple datasets.
			The deeper color, the denser interaction.
			The filters are denoted by a triplet-abbreviation, \textit{i.e.}, “domain-type-kernelsize”, as shown in the vertical and horizontal axes of the up-left sub-figure.
			\textbf{S} and \textbf{T} denotes \textit{spatial} and \textit{temporal} domain,
			\textbf{C}, \textbf{M} and \textbf{I} denotes \textit{convolution}, \textit{max-pooling} and \textit{identity}.
			For example, \textbf{S-C-3} denotes the spatial convolution with $3\times3$ kernel size.
			All the sub-figures share the same axes caption as the up-left one for brevity.
			\textit{Best viewed by zooming in.}
		}
		\label{fig_interaction}
	\end{figure}
	
	To understand how spatial and temporal information are interacted in the \textit{action perceptron synthesizer} to form the optimal kernel,
	we check the weights of the proposed \textit{optimized feature fusion layer} in the synthesizer after each residual block of the ResNet-50 backbone, as shown in Fig.~\ref{fig_interaction}.
	Specifically, we quantize the importance of different shape interactions as the $\ell1$ norm of the corresponding transformation matrix, \textit{i.e.}, $\mathcal{T}_{ij}$ in Eq. (\ref{eq_loss1}).
	We first find that the over-all distribution learned on Something-Something v1 (first row) is non-sparse, indicating the rich-scale spatial-temporal modeling is necessary.
	Moreover, the max-pooling operation occupies considerable importance for serving as a hard attention strategy to find the local key contexts, which is consistent with the quantitative comparison in Tab.~\ref{tab_arch_details}. 
	Also, we find the receptive range of the operation evolves from spatial- to temporal- to spatial-temporal.
	%This is intuitive because the temporal modeling is meaningful only under the premise that the most important objects have been discriminated.
	We further visualize the statistics of the models trained on different datasets.
	For datasets requiring temporal information such as Something-Something v1 (first row in Fig.~\ref{fig_interaction}) and Diving48 (third row in Fig.~\ref{fig_interaction}), we can see the models begin to aggregate the features by $3\times1\times1$ or $5\times1\times1$ temporal convolutions in the first stage, which is not learned out from Kinetics-400 in the early stage.
	Kinetics-400 involves richer spatial-temporal interactions in the second stage (as shown in the second column of the second row in Fig.~\ref{fig_interaction}) for its more diverse background scenes and more complicated object interactions happening in the wild.
	Moreover, for stage 2, 3 and 4, compared to another two datasets, the most interactions of the model trained on Kinetics-400 concentrate on the temporal max-poolings instead of the temporal convolutions,
	%heavily relies on s over the objects with rich spatial scales to be recognized, 
	which aligns with the fact that most action primitives in Kinetics are chronologically irrelevant.

	We demonstrate more analysis \textit{w.r.t.} the most improved classes by our method, action distribution in the feature space, action activation map and the temporal evolution of predictions in the \textit{supplementary material}.

	\section{Conclusion}
	To tackle the essential spatial-temporal scale variances in videos, we propose to learn the optimal-scale kernels from the data and instantiate our method as the action perceptron synthesizer block.
	%	Based on this block, we propose the video perceptron synthesis network for efficient and effective video modeling.
	We perform extensive evaluations to study its effectiveness on video action recognition task, achieving state-of-the-art results.
	% 	on Something-Something V1, Kinetics-400 and Diving48 datasets.
	We also demonstrate some visualization results for more intuitive understanding of our method.

	% or
	

	%\appendices
	%% you can choose not to have a title for an appendix
	%% if you want by leaving the argument blank
	%\section{}
	%% use section* for acknowledgment
	%\section*{Acknowledgment}

	
	\newpage

	
	
	%\bibliographystyle{IEEEtran}
	%\begin{thebibliography}{1}
	%\end{thebibliography}
	\bibliographystyle{aaai2021}
	\bibliography{IEEEfull}

\end{document}